\documentclass[conference]{IEEEtran}
\IEEEoverridecommandlockouts
\usepackage[colorlinks,
linkcolor=red,
anchorcolor=blue,
citecolor=green]{hyperref}

\usepackage{cite}
\usepackage{amsmath,amssymb,amsfonts}
\usepackage{algorithmic}
\usepackage{graphicx}
\usepackage{textcomp}
\usepackage{xcolor}
\usepackage{booktabs} 
\usepackage{array}  
\usepackage{multirow}
\def\BibTeX{{\rm B\kern-.05em{\sc i\kern-.025em b}\kern-.08em
    T\kern-.1667em\lower.7ex\hbox{E}\kern-.125emX}}
\begin{document}

\title{Latent-Info and Low-Dimensional Learning for Human Mesh Recovery and Parallel Optimization}

\author{\IEEEauthorblockN{1\textsuperscript{st} Xiang Zhang}
\IEEEauthorblockA{\textit{School of Information Engineering} \\
\textit{Ningxia University}\\
YinChuan, China \\
zxiang7996@stu.nxu.edu.cn}
\and
\IEEEauthorblockN{2\textsuperscript{nd} Suping Wu*}
\IEEEauthorblockA{\textit{School of Information Engineering} \\
\thanks{* represents corresponding author. This work is supported by the National Natural Science Foundation of China under Grant (62062056), in part by the Ningxia Natural Science Foundation Project under Grant (2024AAC02012) and  the Postgraduate Innovation Project of Ningxia University (CXXM202408).}
\textit{Ningxia University}\\
YinChuan, China\\
pswuu@nxu.edu.cn}
\and
\IEEEauthorblockN{3\textsuperscript{rd} Sheng Yang}
\IEEEauthorblockA{\textit{School of Information Engineering} \\
\textit{Ningxia University}\\
YinChuan, China\\
18791368334@163.com}}

\maketitle

\begin{abstract}
Existing 3D human mesh recovery methods often fail to fully exploit the latent information (e.g., human motion, shape alignment), leading to issues with limb misalignment and insufficient local details in the reconstructed human mesh (especially in complex scenes). Furthermore, the performance improvement gained by modelling mesh vertices and pose node interactions using attention mechanisms comes at a high computational cost. To address these issues, we propose a two-stage network for human mesh recovery based on latent information and low dimensional learning. Specifically, the first stage of the network fully excavates global (e.g., the overall shape alignment) and local (e.g., textures, detail) information from the low and high-frequency components of image features and aggregates this information into a hybrid latent frequency domain feature. This strategy effectively extracts latent information. Subsequently, utilizing extracted hybrid latent frequency domain features collaborates to enhance 2D poses to 3D learning. In the second stage, with the assistance of hybrid latent features, we model the interaction learning between the rough 3D human mesh template and the 3D pose, optimizing the pose and shape of the human mesh. Unlike existing mesh pose interaction methods, we design a low-dimensional mesh pose interaction method through dimensionality reduction and parallel optimization that significantly reduces computational costs without sacrificing reconstruction accuracy. Extensive experimental results on large publicly available datasets indicate superiority compared to the most state-of-the-art.
\end{abstract}


\section{Introduction}
\label{sec:intro}

Recovering 3D human mesh from image and video has broad applications, including interactive games, virtual dress-up, and animation rendering. Recovering 3D human bodies from a single image has made significant progress, but recovering human mesh from video still faces challenges. 

\begin{figure}  
\centering  
\includegraphics[width=0.49\textwidth, height=0.35\textheight]{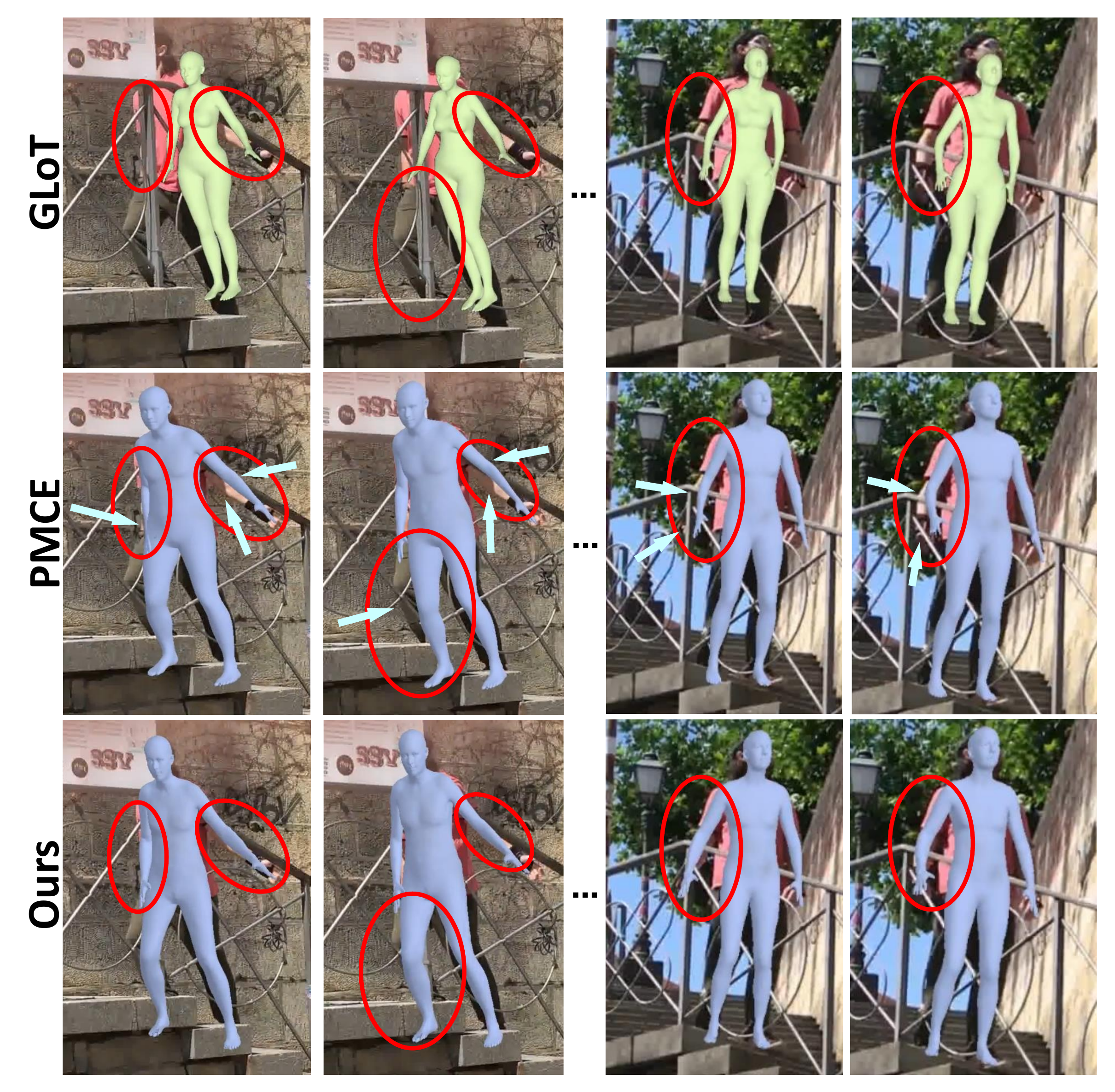}  
\vspace{-0.8cm}\caption{Our method explores the latent information of image features in the frequency domain to enhance representation capabilities. Therefore, despite being in outdoor complex scenes, we can still reconstruct more accurate human mesh. In contrast, the SOTA method easily produces inaccurate human mesh, such as limb location shifts and insufficient local details.}   
\vspace{-0.8cm}  
\label{fig1}  
\end{figure}
Current methods first utilize pre-trained networks like ResNet-50 \cite{He2015DeepRL} for extracting features from input images. Subsequently, a temporal encoder is designed to reinforce spatiotemporal consistency.  GLoT \cite{Shen2023GlobaltoLocalMF} introduced global and local Transformers separate long-term and short-term modeling of image features. To mine more temporal information from the video, Bi-CF \cite{wu2023clip} proposed a bi-level temporal fusion strategy that considers both neighboring and long-range relations. PMCE \cite{You2023CoEvolutionOP} leveraged an attention mechanism guided by temporal image features for mesh-pose interaction, enhancing precise human mesh reconstruction. STAF \cite{yao2024staf} introduced spatiotemporal alignment fusion to preserve spatial information and leverage both temporal and spatial data. UNSPAT \cite{lee2024unspat} proposed a spatiotemporal Transformer to incorporate both spatial and temporal information without compromising spatial information. DiffMesh \cite{zheng2025diffmesh} innovatively connects diffusion models with human motion, resulting in the efficient generation of accurate and smooth output mesh sequences. Although the above methods have made remarkable progress, there are still two issues that need to be addressed. 1) In Fig. \ref{fig1}, current methods often fail to fully exploit the latent information (e.g., human motion, shape alignment), limiting their representative capacity. This can result in inaccuracies such as misaligned limbs and lacking local details in the reconstruction. 2) Current methods commonly utilize attention mechanisms to capture crucial information and feature correlations. However, when the dimensionality of the features is high, using attention mechanisms leads to significant computational overhead. 

\begin{figure*}  
\centering  
\includegraphics[width=1.0\textwidth, height=0.25\textheight]{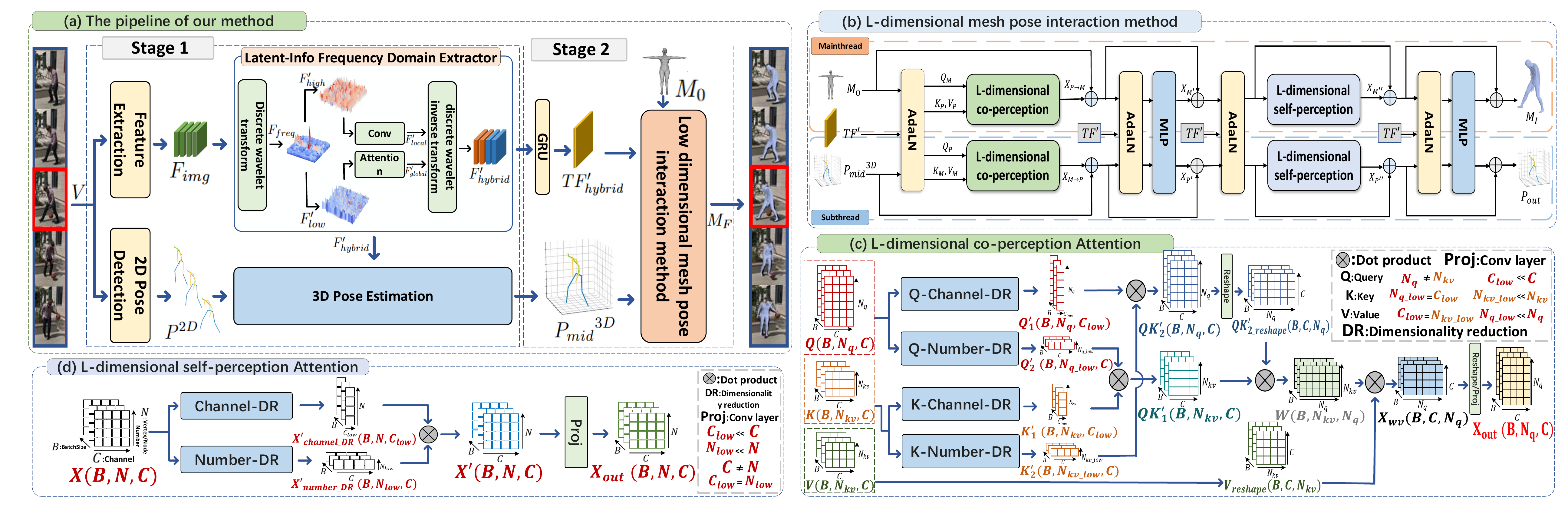}  
\vspace{-1.0cm}\caption{(a) is the framework of our method. Our framework aims to reconstruct the corresponding human mesh given a video sequence. It consists of two stages: latent information extraction and mesh pose interaction modeling. (b) is the architecture of the low dimensional mesh pose interaction method in the stage 2. (c) and (d) are the low-dimensional collaborative perception attention and the low-dimensional self-perception attention in (b), respectively.}   
\vspace{-0.5cm}  
\label{fig2}  
\end{figure*}

To address the issues above, we propose a two-stage network for human mesh recovery based on latent information and low dimensional learning. 
Firstly, we design a latent information frequency domain extractor to fully excavate the latent information. This module decomposes the input image features into low-frequency and high-frequency components, extracting global (e.g., human motion, shape alignment) and local (e.g., textures, detail) information from each. These information features are then aggregated into hybrid latent frequency domain features, which incorporate a contextual understanding of the overall structure extracted from the low-frequency component and an accurate description of the local details captured from the high-frequency component, thereby effectively exploring the latent information of image features and enhancing the representational capacity of the features. These hybrid latent 
frequency domain features provide accurate human shape and motion information to assist 3D pose estimation and mesh-pose co-modelling. Secondly, to reduce the computational cost of mesh vertex and pose co-interaction, we propose a low-dimensional mesh-pose interaction method, using a low-dimensional self-perception attention module, a low-dimensional collaborative perception attention module and parallel optimization.

Our contributions are summarized as follows:
\begin{itemize}
    \item We propose a human mesh recovery network through latent information and low dimensional learning to enhance reconstruction accuracy and decrease computational cost.
    \item We propose a latent information frequency domain extractor to extract latent information effectively from image features. Moreover, our low-dimensional mesh-pose interaction method leverages dimensionality reduction and parallel branching to decrease computational costs without sacrificing reconstruction accuracy.
    \item The experimental results of our method show the effectiveness of the proposed method in widely evaluated benchmark datasets in comparison with state-of-the-arts.
\end{itemize}

\section{Method}

Fig. \ref{fig2} (a) shows the overall workflow of our network, which consists of two stages. The first stage is known as latent information extraction, aimed at extracting hybrid latent features and the 3D pose. The second stage involves modeling the interaction between the mesh and pose to accurately reconstruct a 3D human body mesh. Given an input video sequence $V=\{I_t\}^T_{t=1}$, containing $T$ frames, is processed by a pre-trained ResNet50 to obtain image features $F_{img}\in\mathbb{R}^{T\times2048}$. Simultaneously, an advanced 2D pose estimator provides 2D pose estimates $P^{2D}\in\mathbb{R}^{T\times J\times2}$, where $J$ represents the number of body joints. The 3D pose estimator uses the information provided by hybrid latent features $F'_{hybrid}$ to assist the corresponding 2D pose $P^{2D}$ elevation to 3D space, obtaining the 3D pose ${P_{mid}}^{3D}\in\mathbb{R}^{T\times J\times3}$ as an intermediate representation. Next, the low-dimensional mesh pose interaction method utilizes the temporal features $TF'_{hybrid}\in\mathbb{R}^{2048}$, which are obtained through a GRU, to assist the interaction between the 3D poses ${P_{mid}}^{3D}$ and the initial mesh template $M_0\in\mathbb{R}^{431\times3}$ \cite{Loper2023SMPLAS}, obtaining the intermediate mesh $M_I\in\mathbb{R}^{431\times3}$. Finally, the intermediate mesh $M_I$ is upsampled to a detailed human mesh $M_F\in\mathbb{R}^{6890\times3}$ and outputted. In the following sections, we will provide a detailed explanation of each module.

\subsection{Stage 1:Latent Information Extraction}
\paragraph{Latent Information Frequency Domain Extractor}
The discrete wavelet transform (implemented based on fast algorithms) not only has the characteristic of decomposing signals into different frequency sub-bands, but also has a lower time complexity compared with other frequency domain transform algorithms. Therefore, we apply the discrete wavelet transform to decompose the input image features $F_{img}\in\mathbb{R}^{T\times2048}$ into low-frequency components $F'_{low}$ and high-frequency components $F'_{high}$. 
The mathematical expression is as follows:
\begin{eqnarray}
\begin{array}{lr}
F'_{low} = \sum_{n} F_{img}[n] \cdot \phi(2^{-j}n - k), \\
F'_{high} = \sum_{n} F_{img}[n] \cdot \psi(2^{-j}n - k).
\end{array}
\end{eqnarray}
here, $F_{img}[n]$ denotes the value of the input signal at position $n$. $\phi$ is the scaling function used to extract the low-frequency part of the signal, while $\psi$ is the wavelet function used to extract the high-frequency part of the signal. $2^{-j}$ serves as the scaling factor, representing the decomposition at the j-th level. $n$ denotes the index of the input signal, and $k$ denotes the index of the approximation coefficients. 
The $F'_{high}$ and $F'_{low}$ are then separately fed into different feature extraction branches. To get the global and local information, inspired by \cite{Yun2023SPANetFT}, in the low-frequency part, we use attention that has a strong low-pass filtering capability to get the global information $F'_{global}$ in $F'_{low}$ (like human motion and shape alignment). In the high-frequency part, we use convolutional layers that have excellent high-pass filtering ability to catch the local details $F'_{local}$ in $F'_{high}$ (such as the shape and position of hands and feet). 
\begin{eqnarray}
\begin{array}{lr}
Q = W_Q \cdot F'_{low}, K = W_K \cdot F'_{low},V = W_V \cdot F'_{low}, \\
F'_{global} = \text{Softmax}\left(\frac{QK^T}{\sqrt{d_k}}\right)V, \\
F'_{local} = \text{Conv}(F'_{high}).
\end{array}
\end{eqnarray}
here, $W_{(\cdot)}$ is the weighting matrix, $d_k$ denotes the scaling factor, $Conv$ refers to the convolution operation. We integrate $F'_{global}$ and $F'_{local}$ through discrete wavelet inverse transform to obtain hybrid latent frequency domain features $F'_{hybrid}\in\mathbb{R}^{T\times2048}$ by the following formula. 
\begin{eqnarray}
\begin{array}{lr}
F'_{hybrid} = &\sum_{k} F'_{global}\cdot \phi(2^{-j}n - k) \\ & + \sum_{k} F'_{local} \cdot \psi(2^{-j}n - k)
\end{array}
\end{eqnarray}
where, the functions and variables referred to in Equation 1. $F'_{hybrid}$ contains the fully extracted latent frequency domain information (global and local), thus effectively enhancing the representational ability of the features. 
\paragraph{3D Pose Estimation}
Our 3D human pose estimation adopts the structure of MixSTE \cite{zhang2022mixste}. The input for the 3D human pose estimation consists of 2D pose sequences and corresponding hybrid latent frequency domain features.

\subsection{Stage 2:Mesh Pose Interaction Modeling}

In Fig. \ref{fig2}(b). We propose a low-dimensional mesh pose interaction method (\textbf{LDMP}) that reduces computational cost without sacrificing accuracy. Firstly, we use GRU to extract the temporal information from the hybrid latent features $F'_{hybrid}$, and then we get the temporal latent features $TF'_{hybrid}$, which is used as one of the inputs of LDMP. The LDMP consists of two parallel branches. The mesh branch gradually refines the initial mesh $M_0$ templates using $TF'_{hybrid}$ and 3D poses ${P_{mid}}^{3D}$. The pose branch enhances 3D poses by utilizing $TF'_{hybrid}$ and the human mesh by the following formula.
\begin{eqnarray}
\begin{array}{lr}
M_I = Branch_{mesh}(TF'_{hybrid},M_0,P_{mid}), \\
P_{out} = Branch_{pose}(TF'_{hybrid},P_{mid},M_0).
\end{array}
\end{eqnarray}
where, $M_I$ and $P_{out}$ respectively represent the optimized mesh and pose output. Each branch includes adaptive normalization layers (\textbf{AdaLN}) \cite{huang2017arbitrary} that injects shape information from the $TF'_{hybrid}$ into pose joints and mesh vertex features while preserving spatial structure. MLPs are used for feature aggregation. Since the mesh features and pose features are of very high dimensions, traditional attention methods usually lead to high computational costs when modeling the interactive learning between the mesh and the pose. In fact, there is some redundant information in the mesh and pose features. To solve this problem, we design low-dimensional co-perception attention (LCP) and low-dimensional self-perception attention (LSP) in LDMP. Both of these attention mechanisms first reduce the dimension of the features and then use the low-dimensional features for calculation, to reduce the computational cost. To reduce the computational cost caused by the dimension reduction process, we use simple pooling operations to achieve feature dimension reduction. Next, we will provide a detailed introduction to the LCP and LSP modules.

\paragraph{Low-Dimensional Collaborative-Perception Attention Module}

In Fig. \ref{fig2}(c). To model the interaction learning between the mesh and the pose and reduce the computational cost of this process at the same time, we design the LCP module. We design a novel strategy that pose nodes and mesh vertices are multiplied with learnable weight matrices to obtain Query, Key, and Value as inputs for the LCP. In the mesh branch, we use $Q_{mesh} \in \mathbb{R}^{B\times N_{q}\times C}$, $K_{pose} \in \mathbb{R}^{B\times N_{kv}\times C}$, and $V_{pose} \in \mathbb{R}^{B\times N_{kv}\times C}$ as inputs (in the pose branch, $Q_{pose}$, $K_{mesh}$, and $V_{mesh}$ are used). Here, $B$ is the batch size, $N$ represents the number of vertices/nodes, and $C$ is the channels; $N_{q}$ and $N_{kv}$ are not equal. Initially, dimension reduction is applied to $Q_{mesh}$ and $K_{pose}$ along the dimensions of the number $N$ and channel $C$ via simple adaptive pooling operations. This process yields dimension-reduced outcomes $Q'_{1} \in \mathbb{R}^{B\times N_{q}\times C_{low}}$, $Q'_{2} \in \mathbb{R}^{B\times N_{q\_low}\times C}$ for $Q_{mesh}$ (with $N_{q\_low} \ll N_{q}$ and $C_{low} \ll C$). Similarly, for $K_{pose}$, we acquire $K'_{1} \in \mathbb{R}^{B\times N_{kv}\times C_{low}}$, $K'_{2} \in \mathbb{R}^{B\times N_{kv\_low}\times C}$ after dimension reduction (where $N_{kv\_low} \ll N_{kv}$ and $C_{low} \ll C$). To enable interaction between mesh and pose features, we perform dot product operations between $Q'_{1}$ and $K'_{2}$, as well as $Q'_{2}$ and $K'_{1}$. Subsequently, by applying dot product operations on the adjusted dimension order of the interaction results $QK'_{1} \in \mathbb{R}^{B\times N_{kv}\times C}$ and $QK'_{2\_reshape} \in \mathbb{R}^{B\times C\times N_{q}}$, we get $W \in \mathbb{R}^{B\times N_{kv}\times N_{q}}$. This interaction weight $W$ is then utilized to optimize the mesh features by multiplying with $V_{pose}$. This can be formulated as:
\begin{eqnarray}
\begin{array}{lr}
Q'_{1}=DR_{channel}(Q_{mesh}), Q'_{2}=DR_{number}(Q_{mesh}), \\
K'_{1}=DR_{channel}(K_{pose}), K'_{2}=DR_{number}(K_{pose}), \\
QK'_{2} = Q'_{1} \cdot K'_{2}, QK'_{1} = Q'_{2} \cdot K'_{1}, \\
W = QK'_{1} \cdot QK'_{2\_reshape}, \\
X_{out} = Conv(Reshape(W \cdot V_{pose\_reshape})).
\end{array}
\end{eqnarray}
here, $DR_{channel}$ and $DR_{number}$ are for reducing the channel and number dimensions of the input tensor, $\cdot$ denotes the dot product operation and $X_{out}$ represents the optimized mesh feature. The structure of the LCP in the pose branch and mesh branch is similar, with the only difference being the input and output. The LCP achieves information interaction modeling between mesh and pose by first reducing the dimensions of tensors and then performing dot product operations between mesh and pose tensors, thus reducing computational costs.

\paragraph{Low-Dimensional Self-Perception Attention Module}
In Fig. \ref{fig2}(d). To further optimize the respective internal features of the mesh and the pose after LCP interaction modeling, we designed the LSP module. The process of optimizing the mesh and pose features is completed by two branches respectively. In the mesh branch example, we adopt a ingenious approach. Specifically, mesh features $X$ fed into LSP are split into two branches for dimension reduction using adaptive pooling to decrease channel and vertex dimensions. The process of dimension reduction in LSP is similar to that in LCP. These reduced tensors $X'_{channel\_DR}$ and $X'_{number\_DR}$ from each branch undergoes dot product operation to optimize mesh feature. The output tensors $X'$ from these operations are combined and processed through convolutional layers in mapping modules. This can be formulated as:
\begin{eqnarray}
\begin{array}{lr}
X'_{channel\_DR}=DR_{channel}(X), \\
X'_{number\_DR}=DR_{number}(X), \\
X' = X'_{channel\_DR} \cdot X'_{number\_DR}, X_{out} = Conv(X').
\end{array}
\end{eqnarray}
here, $DR_{channel}$ and $DR_{number}$ are for reducing the channel and number dimensions of the input tensor. The LSP module, by reducing input tensor dimensions before dot product operation, models features efficiently while lowering computational overhead. After undergoing AdaLN and feature aggregation, the LSP produces a rough 3D human mesh $M_I\in\mathbb{R}^{431\times3}$ aligned with the input image, which can be refined into a detailed mesh $M_F\in\mathbb{R}^{6890\times3}$ through upsampling by the following formula.
\begin{eqnarray}
\begin{array}{lr}
M_F=\mathrm{Upsampling}(MLP(AdaLN(M_I,TF'_{hybrid}))).
\end{array}
\end{eqnarray}
The structure of the LSP in the pose branch and mesh branch is similar, with the only difference being the input and output. 

\paragraph{Dual Branch Parallel Computation}
To further reduce time consumption, we introduce an asynchronous parallel strategy to parallelize the LDMP module. The strategy makes use of the thread-level parallel processing ability. The main thread, which can be considered as a group of threads within the CUDA framework, handles the LDMP module's mesh branch calculation. At the same time, the sub-thread is dedicated to the pose branch calculation. Each thread within the respective groups operates independently on its assigned portion of data. They access the GPU's shared memory and registers as required during calculation. With this strategy, the two branches maintain the independence and high efficiency of their calculation processes, accelerating the overall calculation. This can be formulated as:
\begin{eqnarray}
\begin{array}{lr}
M_I = Mainthread(Branch_{mesh}), \\
P_{out} = Subthread(Branch_{pose}).
\end{array}
\end{eqnarray}


\subsection{Loss Function}
Following \cite{You2023CoEvolutionOP,Choi2020Pose2MeshGC}, in the first stage of the training process, we use the 3D joint loss $\mathcal{L}_{joint}$ to supervise the process of 3D pose estimation. In the second stage of the training process, we employ the mesh vertex loss $\mathcal{L}_{mesh}$, 3D joint loss $\mathcal{L}_{joint}$, surface normal loss $\mathcal{L}_{normal}$, and surface edge loss $\mathcal{L}_{edge}$ to constrain the parameter learning of the network.
\begin{eqnarray}
\begin{array}{lr}
\mathcal{L}=\lambda_m\mathcal{L}_{mesh}+\lambda_j\mathcal{L}_{joint}+\lambda_n\mathcal{L}_{normal}+\lambda_e\mathcal{L}_{edge}
\end{array}
\end{eqnarray}
Where $\lambda_m=1$, $\lambda_j=1$, $\lambda_n=0.1$, and $\lambda_e=20$.

\begin{table*}[htbp] 
  \centering
  \caption{Evaluation of SOTA methods on 3DPW, Human3.6M and MPI-INF-3DHP datasets. All methods use pre-trained ResNet50 to extract the features of each frame of the image.}
  \vspace{-0.3cm} 
    \scalebox{0.65}{\begin{tabular}{c|cccc|ccc|ccc|c}
    \bottomrule
    \multirow{2}[4]{*}{Method} & \multicolumn {4}{c|}{3DPW}   & \multicolumn{3}{c|}{Human3.6M} & \multicolumn{3}{c|}{MPI-INF-3DHP} &\multirow{2}[4]{*}{\shortstack{numbe of \\ input frame}} \\
\cmidrule{2-11}         & MPJPE$\downarrow$ & PA-MPJPE$\downarrow$  & MPVPE$\downarrow$ & ACC-ERR$\downarrow$ & MPJPE$\downarrow$ & PA-MPJPE$\downarrow$  & ACC-ERR$\downarrow$ & MPJPE$\downarrow$ & PA-MPJPE$\downarrow$  & ACC-ERR$\downarrow$ &  \\
    \midrule
    Zhang et al. (CVPR 23)\cite{zhang2023two}  & 83.4  & 51.7   & 98.9    & 7.2     & 73.2    & 51.0   & 3.6 & 98.2    & 62.5   & 8.6  &16  \\
    GloT (CVPR 23)\cite{Shen2023GlobaltoLocalMF}  & 84.3  & 50.6   & 96.3    & 6.6     & 67.0    & 46.3   & 3.6 & 93.9    & 61.5   & 7.9  &16  \\
    Bi-CF (ACM MM 23)\cite{wu2023clip}  & 73.4  & 51.9   & 89.8    & 8.8     & 63.9    & 46.1   & \bfseries3.1 & 95.5    & 62.7   & 7.7  &16  \\
    Key2Mesh (CVPR 24)\cite{uguz2024mocap} & 86.7  & 49.8   & 99.5    & -     & 107.1    & 51.0   & 3.4 & - & - & - & 16  \\
    STAF (TCSVT 24)\cite{yao2024staf} & 80.6  & 48.0   & 95.3    & 8.2     & 70.4    & 44.5   & 4.8  & 93.7    & 59.6   & 10.0  & 16\\
    
    PMCE (ICCV 23)\cite{You2023CoEvolutionOP}  & \underline{69.5}  & \underline{46.7}   & \underline{84.8}    & 6.5     & \underline{53.5}    & \underline{37.7}   & \bfseries3.1 & 79.7    & 54.5   & \underline{7.1}  &16  \\
    UNSPAT (WACV 24)\cite{lee2024unspat} & 75.0  & \bfseries45.5   & 90.2    & 6.5     & 58.3    & 41.3   & 3.8  & 94.4    & 60.4   & 9.2  & 16\\
    DiffMesh (WACV 25)\cite{zheng2025diffmesh} & 77.2  & 48.5   & 94.0    & \bfseries{6.3}     & 65.3    & 41.9   & 3.3  & \underline{78.9}   & \underline{54.4}   & \bfseries7.0  & 16\\
    Ours  & \bfseries68.6  & \underline{46.7}   & \bfseries81.9      &\underline{6.4}     &\bfseries50.6     & \bfseries36.4     &\underline{3.2} &\bfseries75.5    &\bfseries 54.1   & 7.3  &16  \\
    \bottomrule
    \end{tabular}}\vspace{-0.5cm}
  \label{tab1}%
\end{table*}%

\section{Experiments}
\subsection{Implementation Details}
According to \cite{You2023CoEvolutionOP}, we set the input sequence length to 16 and use the pre-trained ResNet-50 as the image feature extractor. For 2D pose detection, we adopt the same configuration as described in \cite{You2023CoEvolutionOP}. Both training stages of the network are optimized using Adam. In the first stage, we train the Latent Information Frequency Domain Extractor and the 3D pose estimator with a batch size of 64, a learning rate of $5\times{10}^{-5}$, and a duration of 20 epochs. In the second stage, we load the weights obtained from the first stage and train the entire network with a batch size of 32 and a learning rate of $5\times{10}^{-5}$ for 10 epochs. Our code is implemented in Pytorch, and the training is performed on an NVIDIA RTX 6000 GPU.
\subsection{Evaluation Datasets and Metrics}
Building upon previous approaches\cite{wei2022capturing, You2023CoEvolutionOP}, we utilize a combination of 2D and 3D datasets for training our model. In particular, we utilize the 3D datasets including 3DPW\cite{von2018recovering}, Human3.6M\cite{ionescu2013human3}, and MPI-INF-3DHP\cite{mehta2017monocular} with annotations for 3D joints and SMPL parameters\cite{Loper2023SMPLAS}. For 2D datasets, we incorporate COCO\cite{lin2014microsoft} and MPII\cite{andriluka20142d}. Accuracy metrics include mean joint position error (MPJPE), Procrustes-aligned MPJPE (PA-MPJPE), and mean vertex position error (MPVPE) measuring Euclidean distance between predicted and ground truth vertices. For temporal evaluation, the mean error between predicted 3D coordinates and ground truth acceleration (ACC-ERR) is computed.

\begin{table}[htbp]
  \centering
  \caption{Performance comparison with SOTA video-based methods on 3DPW without using 3DPW training set during training.}
  \vspace{-0.3cm} 
    \scalebox{0.65}{\setlength{\tabcolsep}{1.0mm}\begin{tabular}{c|cccc}
    \toprule
    Mesh Recovery (w/o 3DPW in Train) & \multicolumn{1}{l}{MPJPE$\downarrow$} & \multicolumn{1}{l}{PA-MPJPE$\downarrow$} & \multicolumn{1}{l}{MPVPE$\downarrow$} & \multicolumn{1}{l}{ACC-ERR$\downarrow$}\\
    \midrule
    Pose2Mesh\cite{Choi2020Pose2MeshGC} & 88.9 & 58.3 & 106.3 & 22.6 \\
    GloT\cite{Shen2023GlobaltoLocalMF} & 89.9  & 53.5 & 107.8 & 6.7 \\
    PMCE\cite{You2023CoEvolutionOP} & 81.6  & \underline{52.3} & 99.5 & 6.8 \\
    Bi-CF\cite{wu2023clip} & \underline{78.3}  & 53.7 & \underline{95.6} & 8.6 \\
    DiffMesh\cite{zheng2025diffmesh} & 88.7  & 53.0 & 105.9 & \underline{6.5} \\
    Ours & \bfseries75.7 & \bfseries{51.7} & \bfseries92.3 & \bfseries6.4 \\
    \bottomrule
    \end{tabular}}\vspace{-0.5cm}
  \label{tab2}%
\end{table}%

\begin{figure}  
\centering  
\includegraphics[width=0.49\textwidth, height=0.3\textheight]{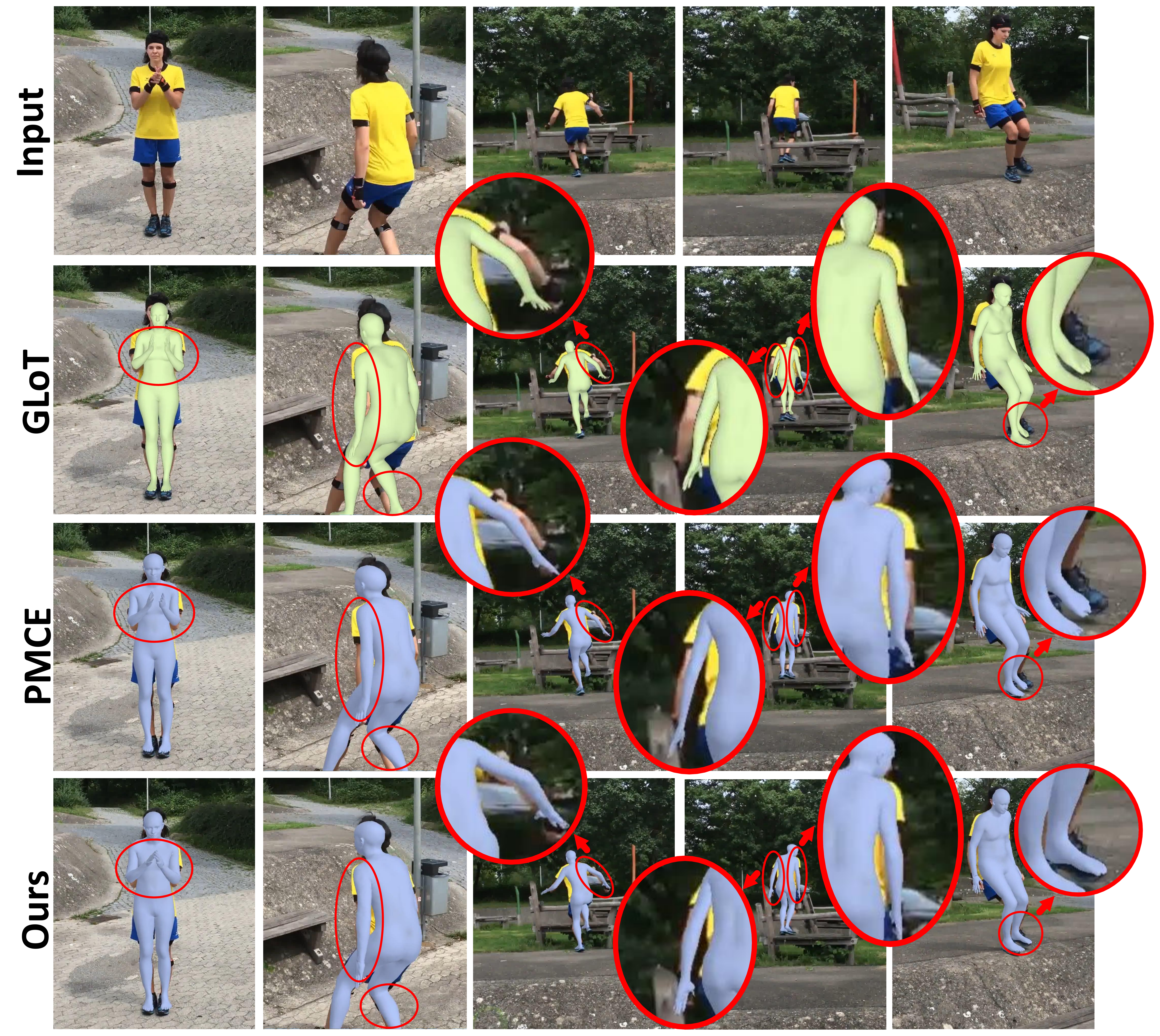} 
\vspace{-0.7cm}\caption{In outdoor complex motion scenes, our method can recover accurate human mesh compared to the SOTA method.}   
\vspace{-0.7cm}  
\label{fig70}  
\end{figure}

\subsection{Comparison Result and Ablation Study}
\paragraph{Comparison Result} As shown in Table \ref{tab1}, we report the results of our model on popular HMR benchmarks: 3DPW, MPI-INF-3DHP, and Human3.6M. For a fair comparison, all methods use ResNet-50 to extract the features of each frame of the image. Our method outperforms the previous video-based methods for per-frame 3D mesh accuracy. Specifically, compared to the previous state-of-the-art method PMCE \cite{You2023CoEvolutionOP}, our model achieves a reduction of 0.9$mm$, 2.9$mm$, and 4.2$mm$ in MPJPE metric on 3DPW, Human3.6M, and MPI-INF-3DHP datasets, respectively. Although the performance of PMCE is superior to that of the latest methods on the 3DPW and Human3.6M datasets, its computational cost is quite high because it uses the traditional attention mechanism to conduct interactive modeling on mesh vertices and pose nodes. Unlike PMCE, our method effectively reduces the computational cost through the LCP and LSP modules. Despite UNSPAT \cite{lee2024unspat} having slightly lower PA-MPJPE than ours on the 3DPW dataset, this performance on other error metrics and datasets is much higher. In spite of DiffMesh achieving a marginally better Accel than ours by 0.1 and 0.3$mm/s^{2}$ on 3DPW and MPI-INF-3DHP dataset, it is designed with the aim of improving the smoothness between frames and its estimation error MPJPE is significantly higher than ours by 8.6$mm$ and 3.4$mm$. 
As shown in Table \ref{tab2}, despite not training on the 3DPW dataset for outdoor scenes, our method shows excellent performance. It outperforms PMCE on all metrics. In particular, it reduces by 5.9 $mm$ and 7.2 $mm$ in terms of MPJPE and MPVPE respectively, underlining its robustness in complex outdoor scenarios. Qualitative experiments on the challenging 3DPW dataset are depicted in Fig. \ref{fig1} and Fig. \ref{fig70}, showing our method's effectiveness over GLoT and PMCE. The red circles highlight areas where our approach excels. Our method accurately captures the human body's shape, pose, and local details in complex scenarios by leveraging the latent information frequency domain extractor, which enhances feature representation capabilities by extracting latent information from image features in the frequency domain.

\begin{figure}  
\centering  
\includegraphics[width=0.5\textwidth, height=0.12\textheight]{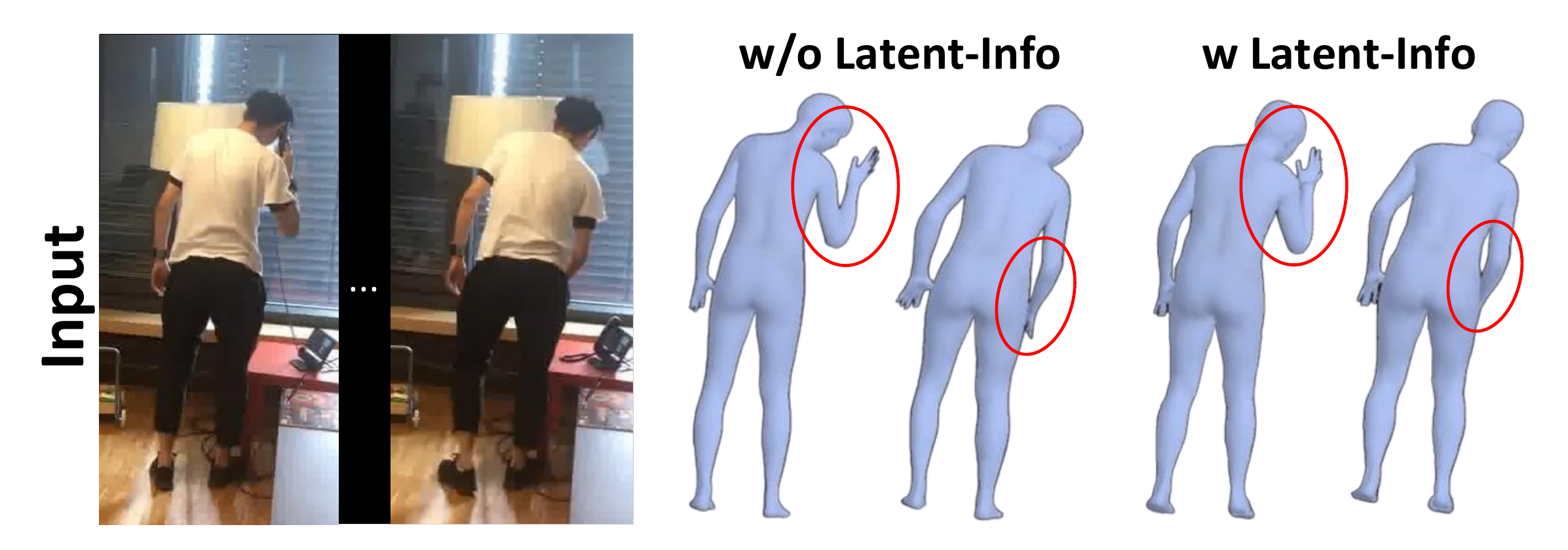} 
\vspace{-0.8cm}\caption{Visual ablation experiments on whether to extract latent information in complex scenes with indoor chaotic backgrounds.}   
\vspace{-0.7cm}  
\label{fig7}  
\end{figure}



\begin{table}[htbp]
\centering
\setlength{\tabcolsep}{4mm}{

\caption{Ablation study for different components.}
\vspace{-0.3cm} 
\renewcommand{\arraystretch}{1.1}
\scalebox{0.65}{\begin{tabular}{c|cccc} 
\specialrule{0.125em}{4pt}{0.1pt}

\textbf{Mesh Recovery(3DPW)}  & \multicolumn{1}{l}{MPJPE$\downarrow$} & \multicolumn{1}{l}{PA-MPJPE$\downarrow$} & \multicolumn{1}{l}{MPVPE$\downarrow$} & \multicolumn{1}{l}{ACC-ERR$\downarrow$}\\
    \midrule
  \hline  
    Ours w/o LIFD-Extractor(HB)     & 77.3 & 52.0 & 94.6  & 6.6   \\
    Ours w/o LIFD-Extractor(LB)     & 77.5 & 52.2 & 94.1 & 6.5   \\
    Ours w/o LDMP              &  76.7 & 51.9 & 93.5 & 6.5 \\
    Ours                      & \bfseries 75.7 & \bfseries 51.7 & \bfseries 92.3 & \bfseries 6.4 \\
  \specialrule{0.125em}{0.3pt}{0.1pt}
\end{tabular}}\vspace{-0.3cm}\label{tab10}
}
\end{table}

\begin{table}[htbp]
  \centering
  \caption{Ablation experiments on whether to adopt the LIFD-Extractor for the 3D pose estimation flow.}
  \vspace{-0.3cm} 
  \scalebox{0.65}{
    \begin{tabular}{c|ccc|ccc}
    \toprule
    \multirow{2}{*}{Method} & \multicolumn{3}{c|}{3DPW} & \multicolumn{3}{c}{Human3.6M} \\
    \cmidrule(lr){2-4} \cmidrule(lr){5-7}
    & MPJPE$\downarrow$ & PA-MPJPE$\downarrow$ & ACC-ERR$\downarrow$ & MPJPE$\downarrow$ & PA-MPJPE$\downarrow$ & ACC-ERR$\downarrow$ \\
    \midrule
    w/o LIFD-Extractor & 70.5 & 44.8 & 6.6 & 52.37 & 37.5 & 3.1 \\
    w LIFD-Extractor & \textbf{68} & \textbf{44.7} & \textbf{6.4} & \textbf{50.9} & \textbf{36.8} & \textbf{3} \\
    \bottomrule
    \end{tabular}
  }
  \label{tab11}%
  \vspace{-0.5cm}
\end{table}

\paragraph{Ablation Study}
In Table \ref{tab10}, ablation studies were conducted on different branches in the latent information frequency domain extractor (LIFD-Extractor). Using high-frequency branch (HB) or low-frequency branch (LB) filters independently led to increased accuracy and acceleration errors, indicating the limited feature extraction capabilities of individual branch. Replacing the LDMP with attention-based methods resulted in inferior performance, highlighting the efficacy of our approach in reducing computational overhead and enhancing reconstruction accuracy. In Table \ref{tab11}, we show the impact of whether to use the LIFD-Extractor on 3D human pose estimation in the first stage of the network. On the 3DPW or Human3.6M datasets, after using the LIFD-Extractor, the intra-frame accuracy and inter-frame consistency of the 3D human pose estimation flow are both improved. This shows the effectiveness of this module. In Fig. \ref{fig7}, we show the visual ablation experiments on whether to apply the LIFD-Extractor in complex scenes with indoor chaotic backgrounds. When we don't use the LIFD-Extractor, the reconstructed mesh of the arm is likely to have the problem of position deviation. The input of latent information helps to improve this problem.

\begin{figure}  
\centering  
\includegraphics[width=0.49\textwidth, height=0.14\textheight]{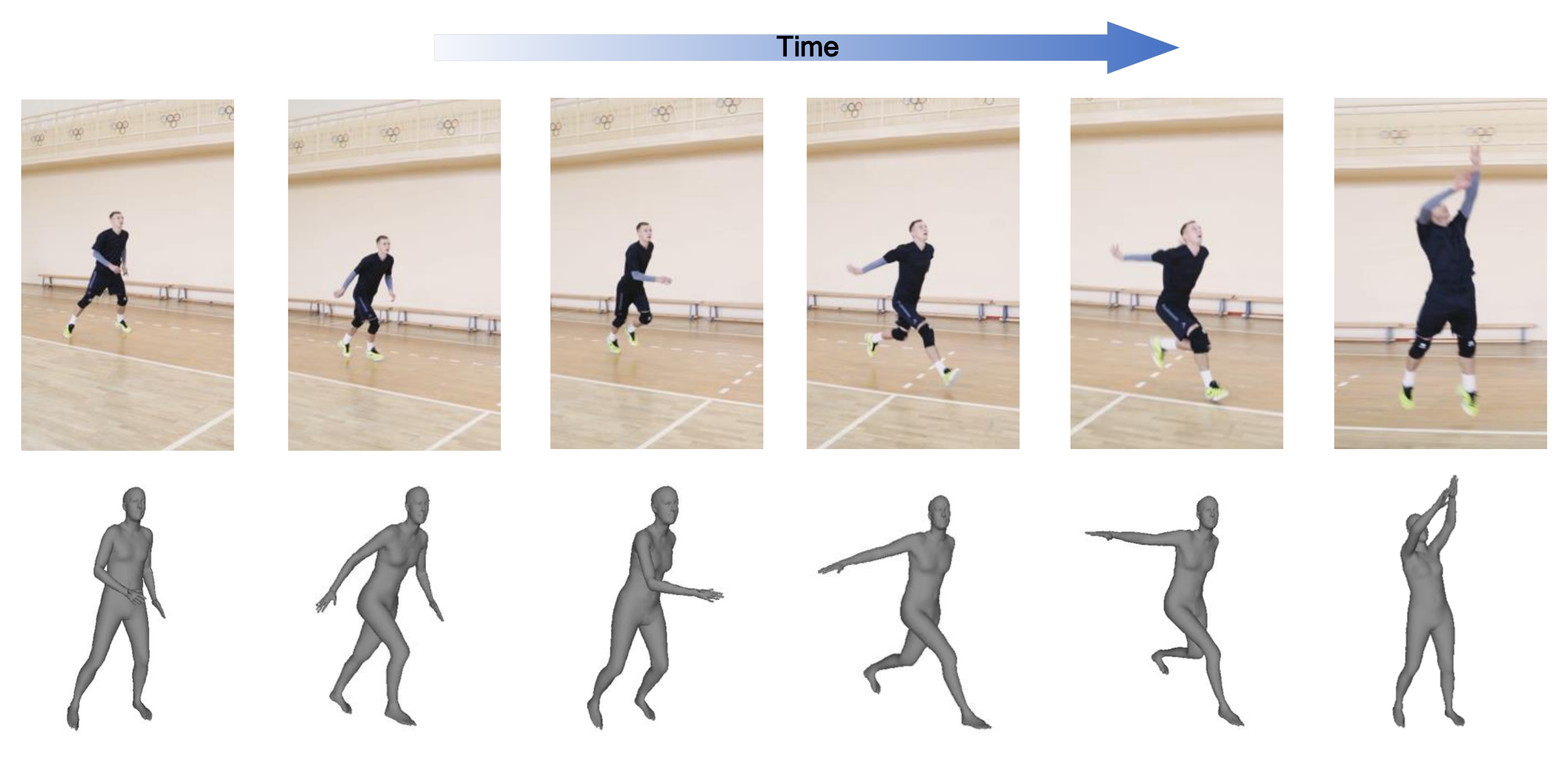}   
\includegraphics[width=0.49\textwidth, height=0.13\textheight]{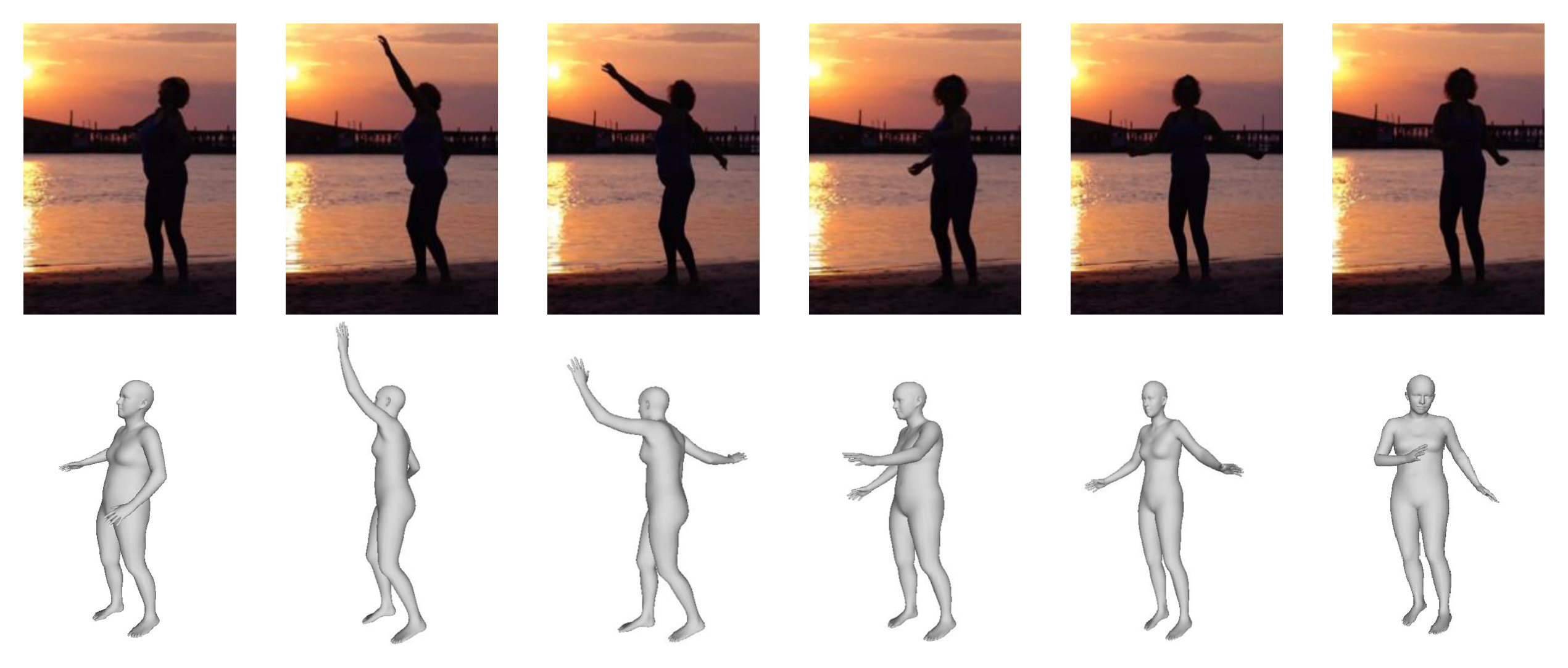} 
\vspace{-0.9cm}\caption{For videos found online, our method can restore accurate and smooth human sequences.}   
\vspace{-0.6cm}  
\label{fig8}  
\end{figure}

\paragraph{Qualitative Results in Generalization.}
To show our method's generalization, we randomly downloaded two videos (with fast motion and complex lighting) from the Internet (see Fig. \ref{fig8}). Then we recovered smooth and accurate human sequences that had better continuity and motion discrimination, which shows our method can make good use of latent information.
\begin{table}[htbp]
\centering
\setlength{\tabcolsep}{4mm}{
\caption{Comparing the parameter count and computational cost.}
\vspace{-0.3cm} 
\renewcommand{\arraystretch}{1.1}
\scalebox{0.6}{\begin{tabular}{c|c|cc} 
\specialrule{0.125em}{4pt}{0.1pt}

  \textbf{Method}  & \textbf{MACs}$\downarrow$& \textbf{Params}$\downarrow$ \\
  \hline  
    joint-SelfAttention    & 46.66M  & 574.27K   \\
    joint-LSP     & \bfseries37.11M(-21\%) & \bfseries558.27K   \\
   \hline  
   joint-CrossAttention   & 171.44M & 1125.37K   \\
    joint-LCP     & \bfseries54.81M(-69\%) & \bfseries1108.99K   \\
   \hline  
   vertex-SelfAttention    & 711.42M & 574.27K   \\
    vertex-LSP     & \bfseries477.61M(-33\%) & \bfseries558.27K   \\
   \hline  
    vertex-CrossAttention      & 617.98M & 1125.37K   \\
    vertex-LCP              & \bfseries494.42M(-20\%) & \bfseries1108.99K \\
    \hline  
   Coevoblock    & 4.74G   & 10.06M   \\
    LDMP    & \bfseries3.36G(-30\%) & \bfseries9.87M   \\
  \specialrule{0.125em}{0.3pt}{0.1pt}
\end{tabular}}


\renewcommand{\arraystretch}{1.1}
\scalebox{0.6}{\begin{tabular}{c|c|c|c} 
\specialrule{0.125em}{4pt}{0.1pt}

  \textbf{LDMP}  & \textbf{Sequential}& \textbf{Parallel}
  & \textbf{Speedup}\\
   \hline  
    Time cost/ms    & 7.515   & 6.592 & 1.14 \\
  \specialrule{0.125em}{0.3pt}{0.1pt}
\end{tabular}}

\renewcommand{\arraystretch}{1.1}
\scalebox{0.6}{\begin{tabular}{c|c|cc} 
\specialrule{0.125em}{4pt}{0.1pt}

  \textbf{Method}  & \textbf{GPU Memory-Usage (MB)}$\downarrow$& \textbf{Training Time /epoch (s)}$\downarrow$ \\
   \hline  
    PMCE\cite{You2023CoEvolutionOP}   & 5719   & 5704    \\
    \hline  
    Ours    & \bfseries5327 & \bfseries5068   \\
  \specialrule{0.125em}{0.3pt}{0.1pt}
\end{tabular}}

\vspace{-0.6cm} \label{tab19}
}
\end{table}
\subsection{Computation Cost and Parameter Comparison}
We use MACs to measure computational complexity because MACs calculate the number of multiplication and accumulation operations in the model, making it a concept that closely reflects actual hardware operations. In Table \ref{tab19}, our LDMP reduces computational overhead by 30\% compared to Coevoblock \cite{You2023CoEvolutionOP}. The LSP in our model decreases computational load by 21\% and 33\% during pose (joint-LSP) and mesh (vertex-LSP) self-perception. Additionally, our LCP reduces computational overhead by 69\% and 20\% in pose-mesh (joint-LCP) and mesh-pose (vertex-LCP) interactions, respectively. To ensure the accuracy of testing time, the following experimental results are averaged over 20 test runs. After parallelizing the computation of the pose and mesh branches in LDMP, we achieved a speedup ratio of 1.14. Our method uses less GPU memory and has a shorter average training time per epoch compared to PMCE. The total number of epochs required for our model to converge is also lower than PMCE, making the training cost of our model lower.



\section{Conclusion}
We propose a human mesh recovery network utilizing latent information exploration and low-dimensional perception. Firstly, in order to fully extract the potential information in the image features, we design an innovative latent information frequency domain extractor. In addition, Our LDMP through dimensionality reduction and parallel optimization significantly reduces computational costs without sacrificing reconstruction accuracy. Quantitative and qualitative evaluations show that our method outperforms the most state-of-the-arts.

\bibliographystyle{IEEEbib}
\bibliography{icme2025references}

\vspace{12pt}

\end{document}